# Proximal Reliability Optimization for Reinforcement Learning

Narendra Patwardhan, Zequn Wang[*],

*Department of Mechanical Engineering- Engineering Mechanics, Michigan Technological University, Houghton MI 49931, USA*

**Abstract**

Despite the numerous advances, reinforcement learning remains away from widespread acceptance for autonomous controller design as compared to classical methods due to lack of ability to effectively tackle the reality gap. The reliance on absolute or deterministic reward as a metric for optimization process renders reinforcement learning highly susceptible to changes in problem dynamics. We introduce a novel framework that effectively quantizes the uncertainty of the design space and induces robustness in controllers by switching to a reliability-based optimization routine. The data efficiency of the method is maintained to match reward based optimization methods by employing a model-based approach. We prove the stability of learned neuro-controllers in both static and dynamic environments on classical reinforcement learning tasks such as Cart Pole balancing and Inverted Pendulum.

## 1. Introduction

In recent years, reinforcement learning has seen incremental growth in replacing classical dynamic programming in the field of control engineering due to it making limited to no assumptions about the dynamics of the system. Instead, it depends upon universal approximating capabilities of the control structure to develop a good control function through trial and error experimentation. The challenge of this approach is to efficiently carry out the exploration, which allows the controller to adapt to a control strategy with satisfactory global performance. We can envision the implausibility of directly employing reinforcement learning approach in designing a controller for a physical system, as the controller may crash during thousands or even tens of thousands of trials needed before it finds a stable control function, thereby making it an impractical practice for designing robust controllers. Since conducting trials, in reality, is often infeasible, usually, a mathematical model of the physical system is constructed in the form of a simulator, the controller is designed for the model, and then the controller is implemented on the physical system. If there are substantial differences between the model and the physical system, often called the reality gap, then the controller may operate with compromised performance and possibly be unstable. Physical systems often possess underlying dynamics that are difficult to measure accurately such as friction, density distribution, and unknown torques. Furthermore, the dynamics of the system often change over time; the change can be gradual such as when devices wear or new systems break-in or the change can be abrupt as in the catastrophic failure of a sub-component or the replacement of an old part with a new one.

---

Physical systems as opposed to simulation environments, also rarely provide complete state information. Since state information is extracted in the form of external observations, the problem of state estimation presents itself in a couple of major ways. The first being partial observability, depending on the sensing elements mounted on the robotic agent, there could be some states which cannot be observed from the data gathered, and the second one is the inherent noise associated with each of the sensing element. While for a specific robotic task, the constraint of partial observability can be safely ignored, the inherent sensing noise, which differs from sensor to sensor and the conditions of sense can be significant. The control of a robotic agent is usually done using an electric or mechanical form of signal. In case of partial or minor system failures, the control signal can be expected to deviate from the intended value as determined by the policy structure. In normal working conditions, the deviation is insignificant in magnitude for electric signals and is therefore generally ignored, however for mechanical systems it could be enough to have an effect on policy performance.

Reinforcement learning also faces the problem of test generalization, as the simulation models are often created and evaluated in the same deterministic environmental scenarios. However, the real world, environmental conditions such as brightness, humidity, viscous drag do not stay constant with temporal and locational changes. It is often hard, therefore, for a controller designed in a particular environment to be equally applicable to a similar environment with minor changes in ambient conditions. This combined form of dynamism is especially hard to capture even with high fidelity simulator, which imposes progressively severe overhead in computational time with an increase in fidelity levels.

In this paper, we alleviate this dependency on modeling fidelity and extensive prior knowledge requirement by providing a framework that tackles various sources of uncertainty by taking the effect of systematic variability in account while optimizing as opposed to a single deterministic reward. To provide a comparable metric to methods established in the classical control literature, that provides assurance of controller under uncertainty, we utilize the well-established concept of reliability from the domain of engineering design. Reliability is defined as the probability that system performance meets its marginal value while taking into account the uncertainties involved in the system at the run time. By embedding the reliability information in the optimization procedure, we propose a novel Proximal Reliability Optimization-based Reinforcement Learning (PRO-RL) to achieve policies which exhibit much robust empirical performance against both time-invariant and time variant changes in the governing system parameters.

This article first presents a survey of related work in Sec. 2 and then introduces the virtual environment modeling procedure for improved data efficiency. Sec. 4 delves deeper into the policy optimization routine. We provide a detailed evaluation of the PRO-RL algorithm on two classical reinforcement learning problems reimagined as probabilistic environments in Sec. 5. Sec. 6 concludes the paper with key results of PRO-RL algorithm.

## 2. Related Work

One of the primary techniques established among the prior attempts to handle the uncertainty associated with reinforcement learning in the domain of robotics is that of dynamics randomization [1], which approaches the problem of generalization from the completely stochastic perspective, changing the physical settings of the experiment after each run. This, in turn, avoids the over-fitting of the controller to a specific set of physical conditions, however, does not provide a guarantee that policy obtained from the optimization procedure is bound to work for entire design space or to be equally effective for the scenarios sampled.

Another related technique, domain randomization, is presented in [2] and [3]. Instead of a high fidelity deterministic simulation, by exposing the policy to a variety of scenarios, domain randomization has been used to bridge the reality gap for tasks such as object localization and robotic grasping. The key contribution of this technique is to provide an alternative for photorealism in simulation for image-based reinforcement learning. While domain randomization is certainly a powerful technique, it is predominantly affected by the amount of training data, thus requiring a large number of sample environments to be created to be effective.

Iterative learning control (ILC), a more data efficient method, employs real-world data from where the system is intended to perform to improve the internal predictive model used to determine controller behavior in the offline stage. Iterative learning control needs prior knowledge in the form of a low fidelity dynamics model, which is used to form the controller for the real system and then closes the loop by gathering the resultant data to improve the dynamics model. Iterative learning control has been applied to a variety of robotic control problems such as robotic arm manipulation by [4], [5], and for Hexapod locomotion in [6] the recent literature. ILC approaches the minimization of the reality gap in a systematic manner but fails to account for the variability in the real world, thereby limiting generalization.

Robust adversarial reinforcement learning or RARL [7] reformulates the problem as a two player game, where the aim of the primary policy is to maximize reward in presence of an adversary which tries to change the trajectory of an experiment by applying available force. Therefore the adversarial policy in RARL imitates the combination of physical and control uncertainties. One of the primary drawbacks of RARL is the memory complexity as it needs to train as well as store the parameters of two distinct policies.

The closest approach to our method is EPOpt [8] which uses parametrized environments and with the goal to maximize reward expectation over models in the source domain distribution. By performing optimization on the subset containing low performing trajectories, the overall performance of the policy is progressively improved. However, the performance of this method is dependent on the value of $\epsilon$ which determines percentile of reward that acts as a base for trajectory selection. The policy optimization procedure is done largely online which can be considered as another drawback due to reduced data efficiency. EPOpt also only considers the physical uncertainty and not the observational or control uncertainty, unlike our method.

## 3. Virtual Environmental Modelling

We consider the environment to be a dynamic system described as

$$s_{t+1} = f_s(s_t, a_t, M_t, E_t, O_t, C_t) \qquad (1)$$

where at time instant $t$, $s_t$ is the state of the system, $a_t$ is the action taken. In addition to these components included in the common formulation of the problem, we consider $M_t$ which is a vector denoting physical parameters intrinsic to the system, and $E_t$ which is a vector denoting the ambient conditions of the environment. $O_t$ is the observational uncertainty vector that affects each of the sensing element or the state and $C_t$ is the control variance. Together $M_t, E_t, O_t$ and $C_t$ are responsible for changes in the dynamics of the system and are represented jointly by matrix $\phi_t$.

Therefore,

$$s_{t+1} = f_s(s_t, a_t, \phi_t) \qquad (2)$$

While direct policy search methods have seen great success in reinforcement learning, each value in $\phi$ can drastically affect the performance of the policy, it is, therefore, quite costly but required to run simulations at the varied combination of $\phi_t$ to obtain better generalization. We therefore turn towards model based methods for retaining the data efficiency. By treating the state transition function as black box we can use surrogate modelling techniques to create a representative model that can approximate the system response by using a low amount of real data as in [9]. In this paper, we use neural networks for approximating the state transition function due to their scalability with the dimensionality of the problem. After the generation of such a model, we can perform the policy optimization completely offline just by predicting the next state. With the surrogate model, we consider a dynamics independent reward function for the experiment, to be defined as

$$r_t = f_r(s_t, s_{t-1}, a_{t-1}, t) \qquad (3)$$

### 3.1 Model initialization with Latin Hypercube Sampling

Latin hypercube sampling (LHS) is an efficient sampling method which divides each dimension of the environment in the desired number of equally probable intervals. Through each interval only a single sample is taken, thus ensuring a global model that is equally performant for the entire state space and has a high exploration-data ratio. Previous approaches like [9] which utilize models created through random roll-outs, do not follow a strategic approach to generate the roll-outs and have a lower exploration-data ratio, and hence have to perform the high amount of trials.

To create a virtual environment using LHS, at any time instant $t$ the dynamism of the system is considered to be bounded by the joint distribution $\Phi$. Depending upon the problem complexity and cost of simulation, the initial number of samples $N$ is defined. The three components of the environment, state-space $s$, action-space $a$ and, dynamism $\Phi$ are sampled together using LHS to obtain $N$ sets of start points denoted as $S_0$, $A_0$, $\phi_0$ respectively. The simulation is ran for a single time instant at each of the starting points and corresponding next states $S_1$ and rewards $R$ are recorded. This collected data forms the basis for the creation of the virtual environment.

### 3.2 Neural Networks based Modelling

After performing LHS and collecting data using simulations, we use a feed-forward neural network with $S_0, A_0, \phi_0$ as training data and $S_1$ as the label for approximating the state transition function. Recurrent networks have been used to model the state-space for reinforcement learning problems, more recently in [9]. However, they model sequences as opposed to discrete values and hence require roll-outs as training data. While this approach works well for a single deterministic state of system dynamism parameters, it can become computationally quite costly to run complete simulations for enough set of dynamism parameters. In this work, we use a common network structure for state modelling (shown in Fig. 1) as a fully connected network with 2 hidden layers which contain 32 neurons each. For the hidden layers, the activation function is kept as Exponential Linear Unit or ELU [10]. For the output layer, the activation was the identity function. The optimization metric was kept as Huber loss which provides for more robust predictions. Adam optimizer is used in all the cases with $10^{-3}$ as the learning rate and the weight decay as $10^{-4}$. The models for classical problems are trained for 50 epochs, whereas the models for advanced problems are trained for 100 epochs. The batch size is kept as 32 in all the experiments. After training, we obtain a model that can successfully provide a predictive estimate for the next state as

$$\hat{s}_{t+1} = \hat{f}_s(s_t, a_t, \phi_t) \tag{4}$$

The reward calculations are done by directly evaluating the predicted state through the reward function. However, if the reward function follows a complex structure, we can fit another neural network to predict the rewards, using same inputs and with $R$ as the label in a similar way.

The neural network based state transition approximation is much more computationally efficient compared to actual simulation as it only operates in weight space without the need to render a screen at each time step, and does not require running the actual simulation engine. While the actual simulation has limits on parallelization depending upon the hardware available, the neural network approximator can handle several different scenarios at the same time effectively.

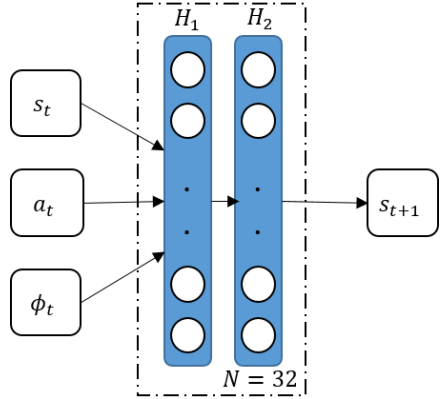

Figure 1: Feed-Forward Neural Network as environment model

## 4. Reliability-based Reinforcement Learning

We consider a stochastic, parametrized policy that maps the state space to action space as $a_t \sim \pi_\theta(\cdot | s_t)$. The first state of the world is sampled from a known distribution $p_0$, which encompasses a subset of the state space.

$$s_0 \sim p_0(\cdot) \tag{5}$$

The physical parameters $\phi_0$ at the start state are kept as $\Phi_{mean}$. For a single realization of dynamism $\phi^i$ we have the realization of the system dynamics as

$$s_{t+1}^i = f_s(s_t, a_t, \phi^i | \phi^i \sim \Phi) \tag{6}$$

The realization of reward for same scenario is then

$$r_t^i = f_r(s_t^i, s_{t-1}, a_{t-1}, t) \quad (7)$$

If the number of realizations $i$ is sufficiently high enough ($10^3 - 10^6$), then we can describe the system dynamics in a probabilistic manner by using a kernel density estimator to convert the discrete observations into a probability distribution.

$$s_{t+1} = argmax_i \, Pr(s_{t+1}^i) \quad (8)$$

In a similar manner reward distribution can be found by applying the reward function on each individual state realization. The reliability at each time instant is then the probability that reward value is greater than a preset threshold.

We define reliability as:

$$rel_t = \Pr(r_t \geq r_{threshold}) \quad (9)$$

Reliability is then returned as the metric for the performance of system as opposed to the reward in deterministic methods at each time instant. The next state is selected as the most probable state from the predictive next state distribution. The entire procedure for per step reliability calculation is highlighted in Fig. 2.

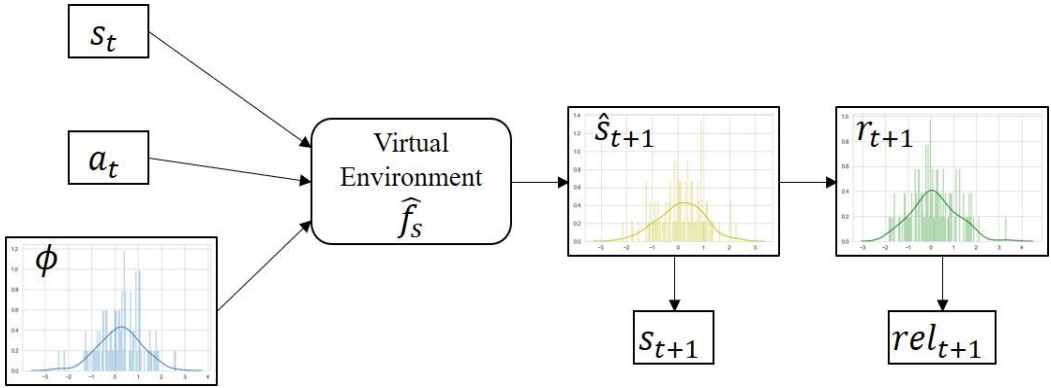

Figure 2: Per Step Reliability Computation using Virtual Environment

The probability of the $T$ step trajectory $\tau$ given policy $\pi$ is

$$\Pr(\tau|\pi) = p_0(s_0) \prod_{t=0}^{T-1} \Pr(s_{t+1}|s_t, a_t) \pi(a_t|s_t) \quad (10)$$

The reliability return of trajectory is given by

$$R(\tau) = \sum_{t=0}^{T} rel_t \quad (11)$$

where $T$ denotes the time horizon that depends upon the problem settings. The expectation of finite horizon reliability returns is given by

$$J(\pi) = \int_\tau \Pr(\tau|\pi) R(\tau) = E[R(\tau)] \quad (12)$$

Our aim is to find a controller which maximizes the return given in Eq. 12.

$$\pi^* = argmax_\pi J(\pi) \tag{13}$$

We redefine the basic functions used in reinforcement learning literature for reliability maximization. The value function $V^\pi(s)$ provides the expected reliability if starting from state $s$ and following policy $\pi$.

$$V^\pi(s) = E_{\tau \sim \pi}[R(\tau)|s_0 = s] \tag{14}$$

The action function $Q^\pi(s, a)$ provides the expected return as specified by Eq. 12 if starting from state $s$ and following policy $\pi$ after taking action $a$.

$$Q^\pi(s, a) = E_{\tau \sim \pi}[R(\tau)|s_0 = s, a_0 = a] \tag{15}$$

The reliability advantage $A^\pi(s, a)$ provides an estimation corresponding to a policy $\pi$ on how much more reliable is it to take a specific action $a$ in state $s$ over randomly selecting an action according to $\pi$.

$$A^\pi(s, a) = Q^\pi(s, a) - V^\pi(s) \tag{16}$$

This formulation allows us to be consistent with prior efforts in the reinforcement learning literature and to re-purpose any methods used for reward maximization for reliability maximization instead.

Non-gradient based approaches for optimization severely confine the number of parameters and the complexity of the policy structure. Most model based methods such as [4], [9] which use analytical gradients or gradient free optimizers, constrain themselves to simpler policies like linear or single hidden layer feed-forward network, thus drastically limiting their ability to scale with problem complexity. Our approach is agnostic about type of optimizer used. In this paper, we focus primarily on gradient based methods for scalability. Due to applicability to both discrete and continuous action space problems, we apply a modified version of proximal policy approximation (PPO) algorithm for policy optimization. However our method can be easily coupled with any of other alternative methods which exclusively work on one specific type of problem, and potentially perform better, such as RAINBOW [11] for discrete action spaces or deep deterministic policy gradients (DDPG) [12] for continuous action spaces.

Policy gradient methods usually suffer from high variance in gradient estimation. Actor-Critic methods overcome this problem by providing a stable baseline. The critic in the actor-critic methods is only used at training time and is used to approximate the value function, which in our case is used to compute the advantage. The actor is the primary structure responsible for determining the course of actions to take in order to maximize the expected cumulative reliability of the agent. Both actor and critic are parametrized using neural networks as shown in Fig. 3. It is possible to reduce the number of parameters and thus the number of trials by using shared weights between actor and critic networks as we are training on synthetic data obtained from virtual environment.

For robustness in training, and to minimize the change per optimization step, PPO maximizes a surrogate objective. Two different approaches for defining the surrogate function are presented in [13], namely, advantage clipping and Kullback–Leibler (KL) penalty. The advantage clipping based approach has been shown to work better empirically and hence is used in our implementation as well. The value function required to compute the advantage is obtained using the flow described in Fig. 4.

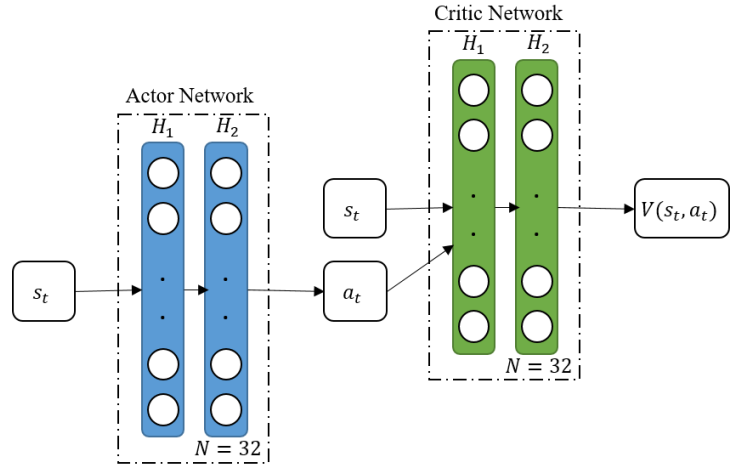

Figure 3: Feed-Forward Neural Networks as Actor-Critic Pair

The new surrogate objective is defined as:
$$L(\theta) = E_t[\min(\rho_t A_t, clip(\rho_r, 1-\epsilon, 1+\epsilon)A_t)] \tag{17}$$

where,
$$\rho_t(\theta) = \frac{\pi_\theta(a_t|s_t)}{\pi_{\theta_{old}}(a_t|s_t)} \tag{18}$$

After obtaining the optimized policy, the performance is validated in the real environment by utilizing desired system dynamism realization which could be completely different from those used to train the predictive model as well as time dependent.

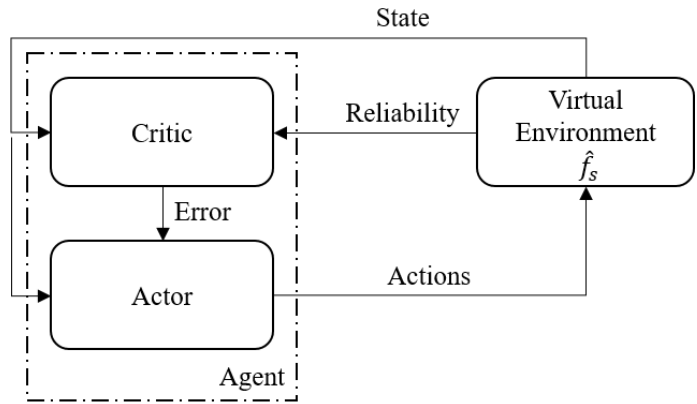

Figure 4: Information Flow in Actor-Critic Architecture

# 5. Experimental Results

In this section, the various design decisions specified in our algorithm are justified by conducting a detailed ablation study on two cases of classical control problems - Inverted Pendulum and Cart-pole Balancing, showcasing the applicability over continuous as well as discrete domain.

## 5.1 Task Descriptions

**Cart-Pole balancing:** The $r_{threshold}$ environment consists a pendulum attached through an un-actuated joint to a cart which moves on a friction-less track. Initially the pole is kept vertical with slight variations in the state parameters drawn from uniform random distribution $U(-0.05, 0.05)$ giving rise to the $p_0(\cdot)$. The aim is to keep the cart-pole balanced between the state constraints for more 195 steps. Each trial in the environment is run for at most 200 time steps, where a single time step is 0.02 seconds. The failure bounds for the position and the angle are $\pm 2.4$ m and $\pm 12°$ respectively.

The reward value is 1 for every successful state and 0 for failure state. If a failure state is observed, all the subsequent rewards are 0. In the deterministic version presented in prior literature, the mass of the pole is 0.1 kg whereas the mass of the cart is 1 kg. The dynamism of cartpole environment is considered to be independent Gaussian in our case. The length of the pole is 1 m. These values act as the mean values for $\Phi$ in our case. The standard deviation value for each of the physical parameter is considered as 0.333 of the mean value. The control parameter is a discrete value (+1 or -1) indicating the direction a constant force of $10\ N$. We consider control uncertainty of $0.1\ N$ as the standard deviation. In addition the environmental factor of gravity $g = 9.8\ m/s^2$ is treated as a random variable with a standard deviation $0.03\ m/s^2$. The observational uncertainty is introduced via adding a scaled Gaussian noise $s = 0.01$ to the real observations.

**Inverted Pendulum:** The inverted pendulum problem consists of a freely swinging pendulum with unit length and mass. The objective is to obtain a controller which can keep the pendulum vertical (at $0°$) with least angular velocity and least amount of effort. Unlike classical interpretations of this problem where the goal is to minimize the weighted quadratic difference to the target state, our reward function considers the action taken to reach the state in consideration as well, making it a more practical orientated composition of the problem. The problem has continuous observation space of dimensionality 3 and a single action as control variable. The nature of action space as well as the rewards are continuous. The states are $[l\cos(\theta), l\sin(\theta), \dot{\theta}]$ with value of $\theta$ normalized between $[-\pi, \pi]$. The angular velocity $(\dot{\theta})$ is limited between $[-8,8]$ and the applicable control (the joint effort) is between $[-2,2]$. We limit our experiment to a maximum of 200 time steps where each time step is 0.05 seconds. We permit the starting state distribution $p_0(\cdot)$ to be corresponding to the entire state space.

The system dynamism is again modeled as combination of independent Gaussian distributions. The physical parameters of mass and length have standard deviation 0.333 of the usual deterministic value. Gravity is modeled similarly to the cartpole problem. We introduce variation in observational uncertainty by changing the value of internal state $\theta$ by adding scaled Gaussian noise with $s = 0.01$. The angular velocity $\dot{\theta}$ is considered to have higher uncertainty in measurement with increased added scaled Gaussian noise $s = 0.1$. The control noise is similarly modeled with $s = 0.1$.

The reward function for the problem is given by:

$$f_r(s_t, a_{t-1}) = -(\theta^2 + 0.1\,\dot{\theta}^2 + 0.001\,a_t^2) \tag{19}$$

Considering the limits on state and action space, the reward is bounded by $[-16.2736, 0]$. For all experiments except where effect of reward threshold ($r_{threshold}$) is studied, the value is set at $-0.01$.

In both the problems, the policy optimized in virtual environment is validated in 100 realizations of the real environment, where $\phi$ varied and starting state is drawn from $p_0(\cdot)$. All the experiments were carried out on workstation with 64 GB RAM.

## 5.2 Performance Evaluation

We evaluate the performance of the learned controller by creating deterministic scenarios by sampling the dynamism parameters $\phi$ in a strategic way from $\Phi$. For the CartPole balancing task, the mass of cart and mass of pole has highest impact on the governing equations. We uniformly sample mass of cart and mass of pole from with limits as three times of standard deviations from mean on either sides. The same experiment is ran for combinations of length of pole and gravity pair. The rewards are presented in Fig. 5. It is clear that throughout the design space variations, the learned controller is reliable to achieve the minimum reward of 195 that is needed to be considered as success.

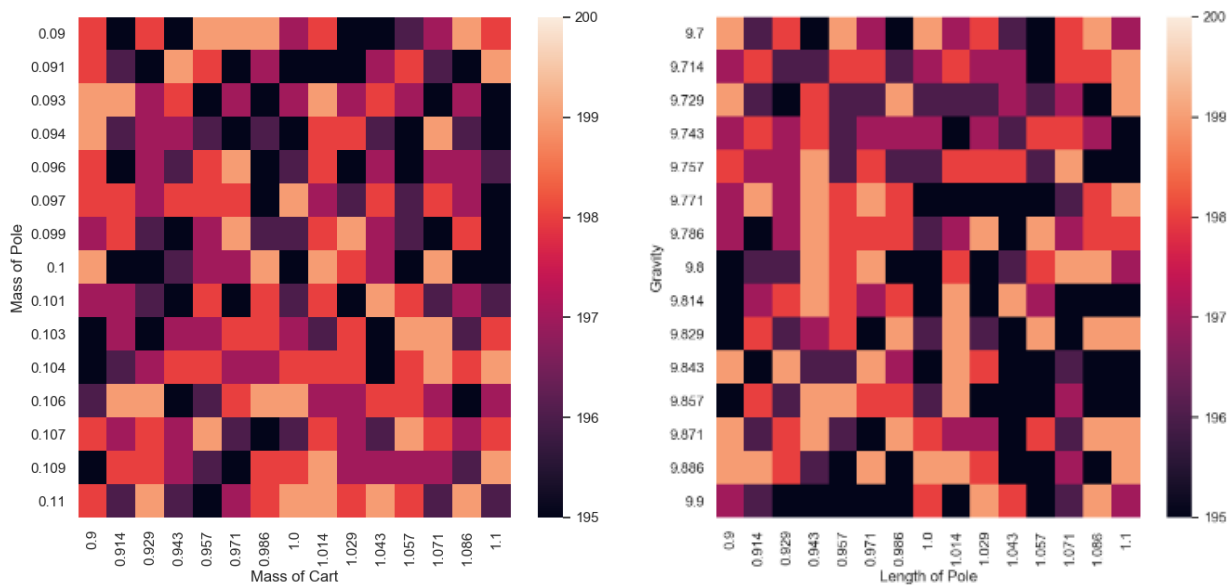

Figure 5: Reward Map for Sampled Realizations of the Pendulum-v0 environment

We evaluate our controller on 1000 random realizations of the environment and present the minimum number of time steps required to attain the position of maximum reward encountered through the trial. Fig. 6 shows that the learned controller generalizes well across the dynamism space and providing equal performance and stabilizes in 100 time steps in majority of cases.

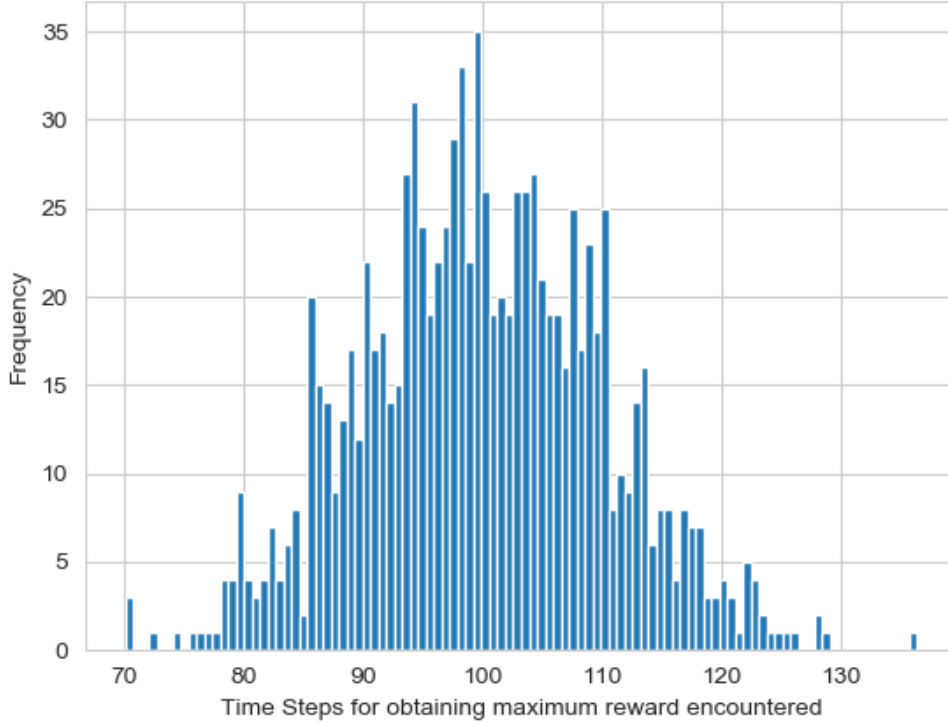

Figure 6: Temporal performance on Sampled Realizations of the Pendulum-v0 environment

## 6. Conclusion

PRO-RL gives a practical solution to autonomously find controllers suitable for use in real environments with bounded uncertainty and eliminates several drawbacks from established reinforcement learning methods such as need for high fidelity simulation, low data efficiency and lack of generalizability in the learned controller.

While many methods consider modelling the dynamics of systems for controller design, they limit their usage of uncertainty information inherent to the system dynamics and focus on a deterministic variant. Our method effectively uses this information for overcoming the reality gap.

Our method is highly able to scale up the complexity of policy structures involved while keeping the computation time much lower than analytical gradient approaches as well as other gradient free optimization methods. Even though the number of queries for virtual environment are orders of magnitude more (depending on the number of realizations used), our computation time is roughly similar to the established model-free method in reinforcement learning that we couple with.

Due to confidence of model being restricted to only previously visited area, most model based methods exploit a small initial state distribution, leading to low robustness. By giving importance to global model fidelity level, by using Latin hypercube sampling method, our method shows increased data efficiency that scales up well with increase in the total problem dimensionality.